\documentclass[letterpaper, 10 pt, conference]{ieeeconf}  

\IEEEoverridecommandlockouts                              

\usepackage{amsmath,amssymb}
\usepackage{cite}
\usepackage{graphicx}
\usepackage{xcolor}
\usepackage{subcaption}
\usepackage{url}
\usepackage{siunitx}
\usepackage{booktabs}
\usepackage{comment}

\title{\LARGE \bf
Bayesian Decentralized Decision-making for Multi-Robot Systems: Sample-efficient  Estimation of Event Rates
}

\author{Gabriel Aguirre, Simay Atasoy Bingöl, Heiko Hamann, and Jonas Kuckling
\thanks{
*JK, SAB, and HH acknowledge support from DFG through Germany's Excellence Strategy-EXC 2117-422037984 and Centre for the Advanced Study of Collective Behaviour (CASCB), Universität Konstanz, Konstanz, Germany.
JK acknowledges support from the Zukunftskolleg Konstanz and the Carl-Zeiss-Foundation. SAB acknowledges support from the Hector Foundation~II.}
\thanks{All authors are with the Dep. of Computer \& Information Science, University of Konstanz, Germany.
        {\tt\footnotesize heiko.hamann@uni-konstanz.de}}%
\thanks{
GA, SAB, and JK conceived the experiments. 
GA performed the experiments.
The paper was drafted by GA, SAB, and JK and edited by GA, SAB, JK, and HH.
All authors read and approved the final version.
The research was directed by JK and HH.
}}

\begin{document}

\maketitle
\thispagestyle{empty}
\pagestyle{empty}

\begin{abstract}
Effective collective decision-making in swarm robotics often requires balancing exploration, communication and individual uncertainty estimation, especially in hazardous environments where direct measurements are limited or costly. We propose a decentralized Bayesian framework that enables a swarm of simple robots to identify the safer of two areas, each characterized by an unknown rate of hazardous events governed by a Poisson process. Robots employ a conjugate prior to gradually predict the times between events and derive confidence estimates to adapt their behavior. Our simulation results show that the robot swarm consistently chooses the correct area while reducing exposure to hazardous events by being sample-efficient. Compared to baseline heuristics, our proposed approach shows better performance in terms of safety and speed of convergence. 
The proposed scenario has potential to extend the current set of benchmarks in collective decision-making and our method has applications in adaptive risk-aware sampling and exploration in hazardous, dynamic environments. 

\end{abstract}

\section{Introduction}
Collective decision-making under uncertainty is a fundamental challenge in multi-robot systems, including domains such as collective perception, environment classification, and spatial consensus~\cite{Ham2018book,ValBraHamDor2016ants,ValFerDor2017FRAI,almansoori2024evolution}. 
Decentralized systems (e.g., robot swarms) operate under strict limitations on sensing, communication, and memory. This constitutes a challenge. 
Instead of sharing/storing complete observation histories, robots must maintain compact model representations of their knowledge.
It is crucial to develop efficient strategies for collective decision-making, especially when observations are sparse, noisy~\cite{chin2023minimalistic}, and gathered from stochastic processes~\cite{khaluf2019neglected}. 
This is typically characterized as a best-of-$n$ problem~\cite{ValFerDor2017FRAI,reina2017model}. 
In robotic applications, best-of-$n$ decision-making arises in several scenarios, such as selecting optimal exploration areas, allocating resources, or identifying critical objectives~\cite{GutCamMon-etal2010, CamGutNou-etal2010BC}.
The emphasis is on how robots disseminate knowledge to progressively establish agreement.

Applications of best-of-$n$ in multi-robot systems may have additional requirements. For example, sampling of option qualities may be risky or the very purpose of the robots' exploration is to assess danger zones in the environment. 
Risks have been considered explicitly in multi-robot systems before~\cite{vanhavermaet2021mcpo,vielfaure2022dora} but not in collective decision-making. 
In this work, we propose a novel scenario where robots need to collectively estimate rates of hazardous events. 
We investigate how a swarm of simple robots can collectively identify the safer of two spatial regions in an environment where hazardous events occur stochastically.
This is a best-of-2 ($n=2$) decision-making problem. Robots collectively agree on the safer area based on few, noisy observations.
Each region generates discrete hazards according to an unknown Poisson process. Robots must form a consensus about which area poses less risk. 
Each robot models the interarrival time of events using a Weibull distribution.
We develop a decentralized Bayesian decision-making framework that combines the robot's own observations and the experiences of its peers. 
Individual confidence estimates, derived from the model's posterior, govern behavioral transitions within a finite state machine that balances exploration and information sharing.
We integrate this with a modified version of the direct modulation of majority-based decisions (DMMD) algorithm~\cite{ValHamDor2015aamas}, which incorporates shared Bayesian parameters. The robot system converges on the correct decision while minimizing exposure. 
Our simulation results demonstrate that this approach outperforms baseline heuristics in both decision accuracy and speed of convergence.

\section{Related work}
Several strategies have been developed that focus on the accuracy, scalability, robustness against uncertainty, and limited communication aspects in collective decision-making~\cite{ValFerDor2017FRAI,ValHamDor2015aamas,jamshidpey2023reducing,khaluf2022robot}. 
Related research addresses spatial decision-making for globally distributed features.  
For example, robots sample locally the frequency of color patches and disseminate their opinions to classify the dominant ground color~\cite{EbeGauNag2018aamas,AusTalDorHamRei2022ants,ZakDorRei2022ants}. 
Consensus building is often based on majority rule (an agent adopts the majority opinion among its peers), sometimes extended with confidence-weighted or probabilistic modulation strategies. 
While these methods perform well in discrete environments with clearly separable features, they become less suitable when robots must infer continuous-valued environmental properties from indirect or sparse observations. 

Several studies address dynamic environments, where option qualities change over time. 
The swarm must remain responsive on the global level and avoid lock-in states. 
Prasetyo et al.~\cite{PraMasFer2019SI} introduce mechanisms, such as stubborn robots~\cite{masi2021robot} and spontaneous opinion switching, to study a best-of-2 problem with abrupt changes in option quality. 
Similarly, Divband Soorati et al.~\cite{SooKroMen-etal2019iros} enable robots to avoid lock-ins through the swarm's plasticity under dynamic conditions. 
Also Pfister and Hamann~\cite{PfiHam2023iros} address a dynamic environment scenario (see below). 
Usually these approaches assume robots can directly sample option qualities, whereas we consider a harder case where robots infer them indirectly from the timing of stochastic events.

There are also other bio-inspired control systems that work more at swarm level and simplify the individual decision mechanism.  
BEECLUST~\cite{SchHam2011bioinspired} employs basic aggregation behaviors based on environmental cues to collectively agree on an aggregation spot. 
Trabattoni et al.~\cite{TraValDor2018ants} present an example of a hybrid control architecture that adds local leadership and dynamic role allocation to increase adaptability. 

Bayesian approaches have been proposed as an alternative to heuristic decision-making strategies. 
These statistical methods enable robots to form compact belief representations and update their estimates incrementally as new evidence is collected. 
Ebert et al.~\cite{EbeGauMallNag2020icra} use decentralized Bayesian inference to solve binary classification tasks. 
Robots combine Bernoulli-distributed observations of discrete environmental features. 
Pfister et al.~\cite{PfiHam2023iros} extended this to dynamic environments by adding statistical change detection methods with robots resetting beliefs when environments shift. 
They depend on option qualities that are easily and immediately sampled. 
Bartashevich and Mostaghim~\cite{BarMos2021SI} use evidence theory to deal with uncertainty and spatial correlations, but their approach does not directly extend to event-driven dynamics.
Other related methods include Bayesian approaches for censored event rates~\cite{follmann1999bayesian} and UCB-methods for contextual bandits~\cite{zhou2020neural}.

\section{Problem Definition}

\begin{figure}
    \centering
    \includegraphics[width=0.85\linewidth]{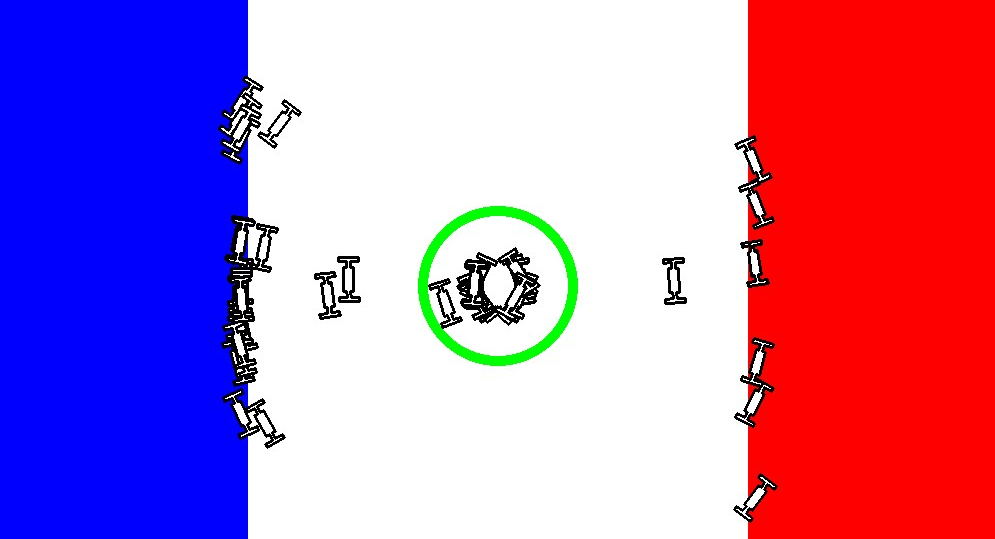}
    \caption{Robot arena (zoomed in): red and blue event areas, central nest zone (green circle), and transition space (white).}
    \label{fig:arena}
\end{figure}

We consider a swarm of $N$ robots operating in an environment divided into four distinct areas, as shown in Fig.~\ref{fig:arena}: event area B(lue), event area R(ed), a central nest (green circle), and a transition zone (white).
We assume that robots are able to accurately move between these zones.
We model hazardous events that occur independently in areas~B and~R by a Poisson process with unknown rates~$\lambda_B$ and~$\lambda_R$, respectively.
While the robots need to minimize their exposure to these potentially dangerous events, we do not actually model any negative consequences resulting from event encounters here (i.e., robots always survive events).
We define the safer area as the one with the lower event rate $\min(\lambda_{B}, \lambda_R)$, that is, where fewer hazardous events occur per time. 
The nest is positioned between area~B and area~R and serves as a zone where robots may communicate with their peers.
The transition zone around the nest and between the two event areas allows robots to move between regions while not allowing event observation or even communication.

Robots have neither prior knowledge of the ground-truth event rates~$\lambda_B$ and~$\lambda_R$, nor can they measure them directly.
Instead, a robot observes the occurrence of an event, when it is in the corresponding event area. 
By moving to areas~B and~R and gathering observations (e.g., measured time between events or between the robot's arrival at the area and the first observed event), robots gradually form and improve beliefs about which area is safer (i.e., rarer events). 
The objective of the swarm is to achieve a consensus on which one is the safer area while minimizing individual exposure to hazardous events (i.e., high sample efficiency).

\section{Method}
We propose a decentralized Bayesian decision-making framework in which each robot builds beliefs regarding the event rates in both areas (B and~R) by observing the time intervals between stochastic (Poisson-distributed) events.
We model these interarrival times using a Weibull distribution to capture non-constant event rates (for details see Sec.~\ref{sec:beliefModel}).
We assume that robots can localize themselves within the environments and distinguish between blue and red regions.
Each robot forms two separate Bayesian models, one for each event area, and uses Bayesian inference to incrementally update its belief about the event rates over time. 
Robots consider estimated event rates and confidence intervals computed from their posterior distributions to minimize the required number of observed events. 
Robots across the swarm collectively form a consensus over event rates driven by the 
Direct Modulation of Majority Decisions (DMMD) algorithm. Their general behavior is determined by a finite-state machine. Robots must determine their observation time, that is, how long to remain in area~B or~R to measure interarrival times. A~promising strategy is to minimize unnecessary exposure by parameterizing the observation time such that a robot observes exactly one event before leaving. Robots must decide their dissemination time, specifying how long they want to inform other robots about their exploration results.

\subsection{Belief model}
\label{sec:beliefModel}
As previously described, the robots need to form a belief about the event rates of areas~B and~R while minimizing their exposure to the events.
The robots model the interarrival time of events, which is inversely related to the event rate.
Whenever a robot observes an event, it computes the interarrival time either since its last observed event or since its arrival to the event area (if no other event was observed since arrival). Our rationale is to increase the number of samples, however, this is at the risk of underestimating the ground-truth interarrival time. We report the results of this estimation strategy in Sec.~\ref{sec:results:time-estimation}.
To minimize memory and communication requirements, robots do not store all observed interarrival times.
Instead, Bayesian inference allows robots to incrementally update their beliefs with minimal data and computational requirements. 
This will allow the same protocol to scale to many-option scenarios without exhausting available memory or communication bandwidth.

We choose the Weibull distribution to model the belief about the interarrival times, as it has been widely used to model time between events~\cite{ShaDicWoo-etal2015}.
The Weibull distribution is parametrized by shape $\beta$ and scale $\theta$. 
The shape parameter~$\beta$ can model dynamic rates, that is, if events are less ($0<\beta<1$) or more ($\beta>1$) likely to happen with increasing time. 
For $\beta=1$ instead, the event rate is constant and the Weibull distribution reduces to an exponential distribution. 
Here, we assume that events follow a Poisson process. Hence, interarrival times of events follow an exponential distribution.
We simplify by setting $\beta=1$ and implement Bayesian inference to estimate the scale parameters~$\theta_B$ and~$\theta_R$. 
We use as conjugate prior a reparametrized inverse-gamma distribution~\cite{Fin1997book}:
\begin{equation}
    \pi(\theta \mid a,b) = \begin{cases} 
    \frac{b^{a-1}\exp(\frac{-b}{\theta})}{\Gamma (a-1)\theta^a}\ & \ \text{where} \ \theta > 0 \\
    0 & \ \text{otherwise}
    \end{cases}\;,
\end{equation}
with gamma function~$\Gamma$, $a$ being the number of encountered events, and $b$ the sum over interarrival times. 
The Bayesian update rules for computing the posterior~\cite{Fin1997book} are
\begin{equation}
    \begin{array}{l l}
        a' = a + n \;,& b'= b + \sum^n_{i=1}x_i \;,
    \end{array}
\end{equation}
with number of observed events~$n$ and interarrival times~$x_i$ of event $i\in\{0,...,n-1\}$. 
As a quantitative model of a robot's uncertainty, we use the $95\%$~confidence intervals of the prior's mode~$D$, which we calculate by $D = {b}/({a + 1})$.

\subsection{Decision algorithms}
\label{sec:decision-algorithms}
During the experiment, the robots need to make assessments and decisions on which event area is safer.
We implemented several strategies that incorporate both the robot's internal beliefs and information communicated by its peers.
Robots start initially in an undecided state.
During the experiment, they gather information and can \textit{select an opinion} on which area is safer.
Robots can change their opinion and even return to the undecided state again.

\noindent
\textbf{Baseline decision algorithm:} We use a baseline for comparison.
There is no inter-robot communication, hence, robots operate in isolation. 
A~robot stays for an initial random time inside the nest before it leaves for an event area. 
When inside the targeted area, it records the number of events it observes. 
A~robot's initial observation time is set by default to $10^3$ time steps. This gets updated based on the number of observations of its latest measurement cycle. 
If the robot experienced more than one event, the timer is decreased by $1\%$ per event observed. 
If it experienced exactly one event, the timer is increased by $1\%$. 
If it experienced no events, the timer is increased by $5\%$. 
The robot chooses uniformly at random which event area to visit (if difference in visits is larger than \num{5}, the less frequently visited one is chosen). 
The robot \textit{selects} an opinion (independent of the area chosen to visit), if each area was visited at least \num{10} times and the robot observed between \num{8} and \num{16} events during its last \num{10} visits. 
We determined these by a preliminary experimental study to avoid biased decision making in early observation periods.

\noindent
\textbf{DMMD:} Our core decision-making approach depends on the DMMD algorithm~\cite{ValFerHamDor2016AAMAS}.
It consists of two phases: dissemination and exploration phase.
In the dissemination phase, robots idle inside the nest and broadcast their current opinion to any neighboring robot for a certain time $t_\text{diss}$ (see Sec.~\ref{sec:times}).
The time a robot remains in the dissemination phase depends on its confidence levels. 
In the exploration phase, a robot explores one of the available areas.
To select an event area, the robot performs a majority vote of its own opinion and all opinions received from neighboring robots during its dissemination phase. 
The robot navigates to that area and measures for an observation time of $t_\text{obs}$ (see Sec.~\ref{sec:times}), before returning to the nest.

We consider the confidence intervals of the mode as indicator about the (un-)certainty of beliefs (see Sec.~\ref{sec:beliefModel}).
If the confidence intervals for the interarrival time estimates of the two event areas are well separated, we consider the difference to be significant and certain. 
Hence, the robot will \textit{select an opinion} corresponding to the event area with the higher of the two interarrival times. 
If the confidence intervals are no longer separated, the robot will retract its opinion and return to an undecided state again.

\noindent
\textbf{DMMD + belief sharing:} 
We need to minimize the number of observations of event areas in addition to achieving consensus quickly. 
Hence, we propose an extension of DMMD that shares also model information in the dissemination phase. 
When any robot switches to the dissemination phase, it shares its own belief, that is, parameters~$a$ and~$b$ while also receiving the beliefs from other disseminating robots. 
It averages all received parameters to $\bar{a}$ and $\bar{b}$ and updates its own parameters: $a' = \frac{a+\bar{a}}{2}$ and $b' = \frac{b+\bar{b}}{2}$.
This allows robots to benefit from observations of other robots, effectively reducing the number of observations required before reaching an informed opinion and minimizing risk. 

\subsection{Robot controller}
Each robot operates a finite-state machine consisting of four states: \textit{Nesting, Leaving, Measuring}, and \textit{Returning}.
Our approach follows the finite-state machine by Valentini~et~al.~\cite{ValHamDor2015aamas}, but our robots do not communicate in the transition zone, but only within the nest.
This simplifies the simulation of communication and ensures that the system is well-mixed (within the nest).
States \textit{Leaving, Measuring} and \textit{Returning} correspond to DMMD's exploration phase and state \textit{Nesting} to the dissemination phase.

\noindent
\textbf{Nesting:} This is the initial state. The robot stays within the nest and disseminates its opinion. When a timer elapses, the robot leaves the nest towards one of the two event areas.
The dissemination time is proportional to the robot's confidence of its belief (see Sec.~\ref{sec:times}).
During dissemination time, robots broadcast the area they believe to be safer and listen to message broadcasts by robots in \textit{Nesting} state.
Once the timer has elapsed, the robot switches to \textit{Leaving} state.

\noindent
\textbf{Leaving:} The robot chooses an area to observe and moves toward the chosen area. 
When entering this state, the robot decides which event area to explore based on a decision algorithm (see Sec.~\ref{sec:decision-algorithms}) that incorporates the robots own beliefs and the information received from other robots during \textit{Nesting} state.
The robot remains in this state for the whole duration of the transition period. 
When it enters the target area, it switches to \textit{Measuring} state.

\noindent
\textbf{Measuring:} The robot remains in the event area and stays for given time to observe events.
The time that a robot waits for events depends on the internal Weibull model of the chosen event area (see Sec.~\ref{sec:times}).
The number of events it can observe in this state is not limited.
After the waiting time has elapsed, the robot updates its internal model (see Sec.~\ref{sec:beliefModel}) and switches to the \textit{Returning} state.

\noindent
\textbf{Returning:} When entering this state, the robot moves toward the nest. 
It remains in \textit{Returning} until it enters the nest and switches to \textit{Nesting} state.

\subsection{Dissemination and observation times}
\label{sec:times}

Robots in \textit{Nesting} or \textit{Measuring} state determine how long to stay based on their belief and uncertainty (see Sec.~\ref{sec:beliefModel}).

\noindent
\textbf{Dissemination time:}
The more confident a robot is about its belief, the more time it will spend disseminating it:
    $\mathit{t_\text{diss}} = \frac{\mathit{c_\text{diss}}}{\mathit{|CI|} + 0.2}$,
where $\mathit{c_\text{diss}}=500$ is a constant and $\mathit{|CI|}$ is the width of the $95\%$ confidence interval. 
We cap the dissemination time at a maximum of  $\mathit{t_\text{diss}^\text{max}}=2500$ time steps.

\noindent
\textbf{Observation time:}
While waiting in the event area, robots aim to observe exactly one event.
We hypothesize that this minimizes both the number of required visits (as robots will never visit an area without observing at least one event) and the number of observed events (as robots rarely observe multiple events during their visit).
To ensure a robot observes at least one event, we use a conservative estimate of the interrarival time. A~robot uses the upper end of the confidence interval $\lceil CI \rceil$ as basis for the waiting time. 
While this increases the likelihood of observation, it also comes with a higher risk due to the longer exposure time. 
The observation time is computed as
    $\mathit{t_\text{obs}} = 2 \lceil CI \rceil$, 
where $2$ is a scaling factor determined by a parameter study.
Although the robot aims to observe exactly one event, it will stay in the event area until the observation time is elapsed, regardless of the number of observed events.

\section{Experimental Setup}
We implemented the simulation in Python using Swarmy,\footnote{\url{https://github.com/tilly111/swarmy}
} a swarm simulator based on pygame.\footnote{\url{https://www.pygame.org/}}
It simulates a swarm of robots with capabilities comparable to  Kilobots~\cite{RubCorNag2014SCI}. 
Our code is available online.\footnote{\url{https://github.com/StudentWorkCPS/gamma_bots}
}
We ran simulations on a single core of an AMD Ryzen~5 7530U CPU at \qty{3.5}{\giga\hertz}. 

We design two environments and assume (w.l.o.g.) that the blue area is safer.
In both environments, we set the rate of the blue area to $\lambda_B=1/(2\times 10^4)$. 
For the red area, the rate is $\lambda_R = 1/10^4$ in the easy environment and $\lambda_R = 1/(1.5 \times 10^4)$ in the difficult environment. 
The reduced difference~$\|\lambda_R-\lambda_B\|$ in rates makes the decision problem more difficult. 


We evaluate four different decision algorithms, formed of combinations of the elements described in Sec.~\ref{sec:decision-algorithms}:
the \textit{baseline} algorithm, with the default observation time initialized as \num{1000} ticks;
the \textit{DMMD} algorithm without belief sharing, with the Bayesian prior initialized as $a=2$ and $b=4\times10^3$;
\textit{DMMD sharing}, where the robots also share their beliefs in addition to their opinion, initialized in the same way;
and \textit{DMMD high prior}, where robots follow the basic DMMD protocol, without belief sharing, but are initialized with a higher prior ($a=2$ and $b=8\times10^4$).
We choose the priors of \textit{DMMD} in such a way that robots initially underestimate the interarrival times of the highest event rate ($\lambda_R$ in the \textit{easy} environment), allowing them to observe few events initially before converging to observing a single event reliably. 
For \textit{DMMD high prior}, we increase $b$ in such a way that robots overestimate initially the interarrival time of all event rates, resulting in observing multiple events before converging to observing only single events.

For all experiments, we set swarm size to $N=50$.
Robots are initialized in the nest with random waiting times, before they explore a randomly chosen side. 
We consider two termination criteria: consensus (\emph{consensus-reaching}) or early termination (\emph{opinion-selection}).
In \textit{consensus-reaching}, we run the experiment until all robots have selected the same opinion.
In early termination, we stop the experiment when all robots have selected an opinion, regardless of whether consensus was reached.
We hypothesize that this could reduce the experiment time while still maintaining high accuracy.
Experiments not finished before \num{10} million time steps are stopped.
Each experiment setup has $64$ repetitions.

Based on preliminary experiments, we implemented three improvements that increased performance and reliability.
In the beginning of the run, robots ignore the majority vote until they have gathered at least eight observations from each side.
This avoids biasing the swarm early-on based on a few outliers.
Additionally, the confidence intervals are computed only from the last 10 observed interarrival times.
This simplifies buffering to a ring-buffer and implements a simple filter that reduces time to commitment when outliers would otherwise dominate the confidence interval for long.
Furthermore, the robots scale down the internally estimated interarrival time by a factor of \num{1000}.
This leads to smaller widths of the confidence interval and therefore more reasonable dissemination times.
See the accompanying video for a run with reduced swarm size and increased event frequency.

\section{Results}

\subsection{Belief model convergence}
\label{sec:results:time-estimation}

\begin{figure*}[tbh]
  \begin{subfigure}{0.3\textwidth}
    \centering
    \includegraphics[width=\linewidth]{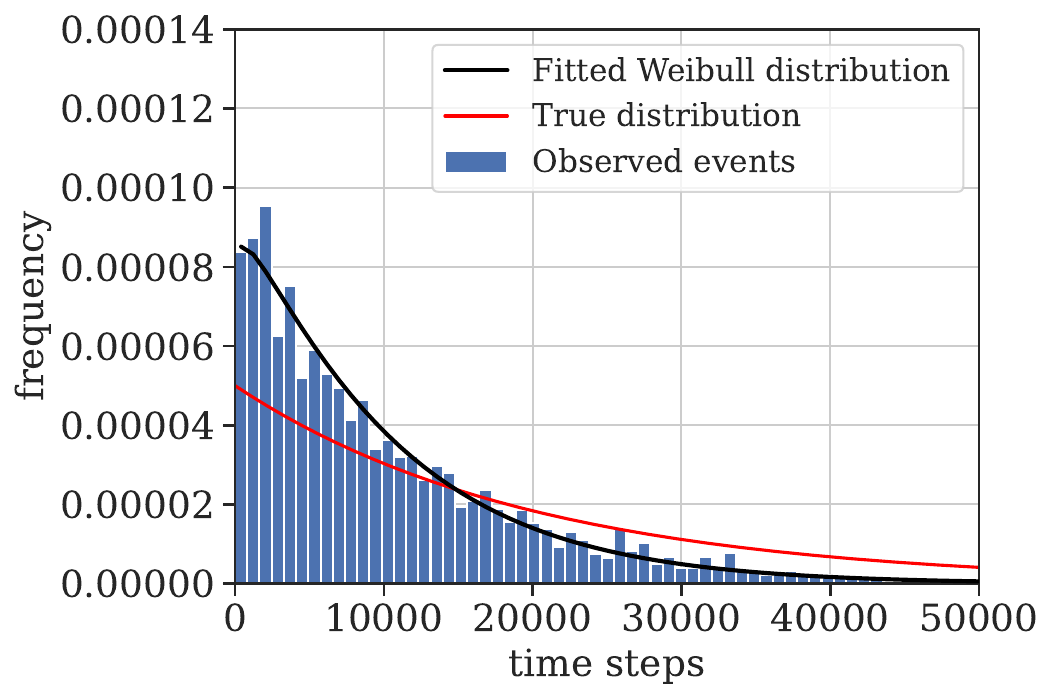}
    \caption{interarrival times, blue area} 
  \end{subfigure}
  \hspace{5mm}
  \begin{subfigure}{0.3\textwidth}
    \includegraphics[width=\linewidth]{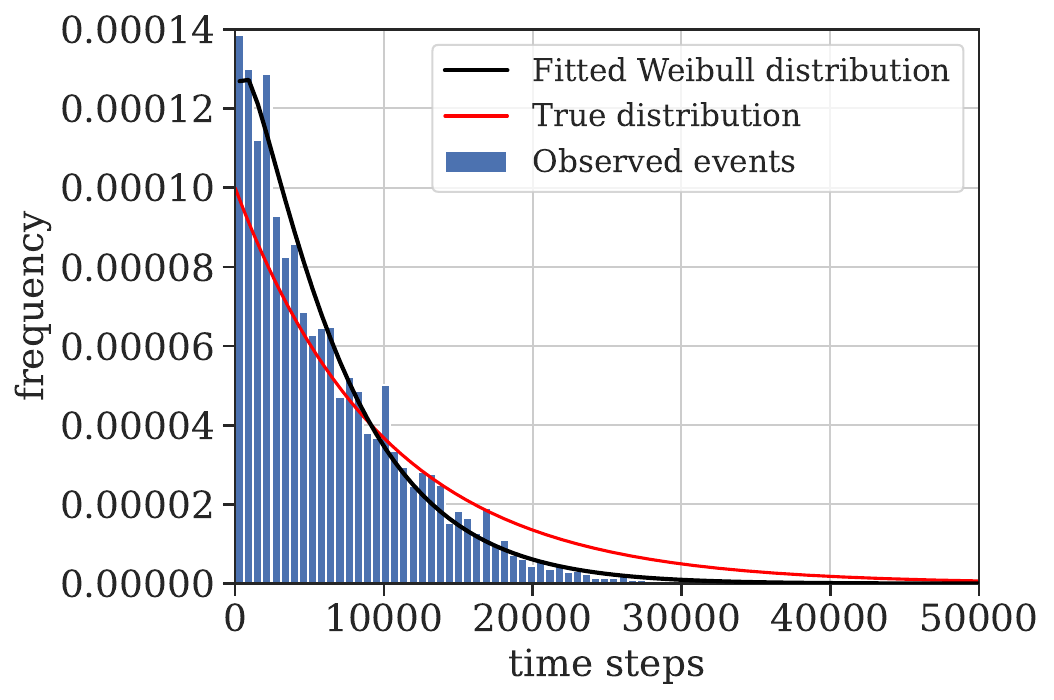}
    \caption{interarrival times, red area}
  \end{subfigure}
  \hspace{5mm}
  \begin{subfigure}{0.33\textwidth}
    \includegraphics[width=\linewidth]{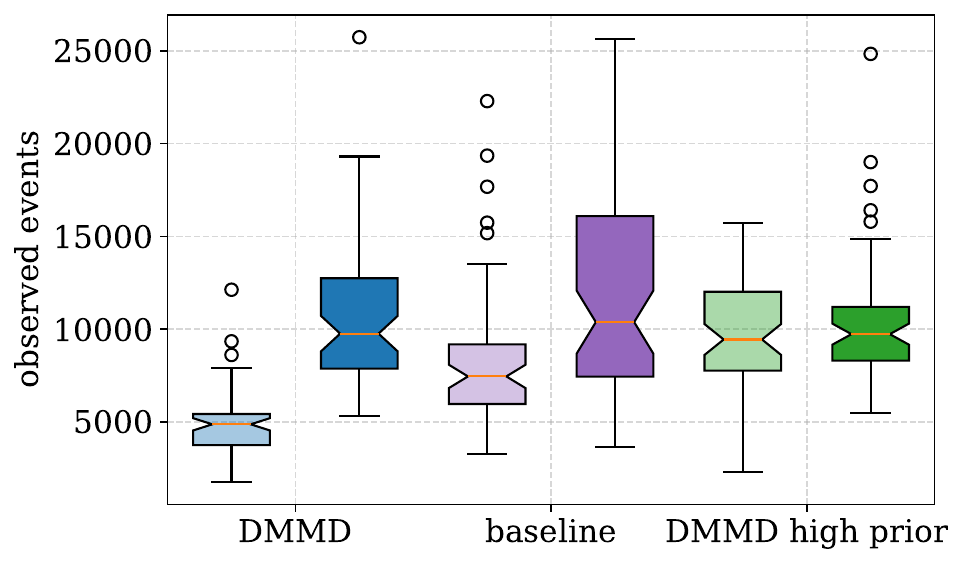}
    \caption{number of observations}
    \label{fig:observed_events}
   \end{subfigure}
  \caption{Histograms of measured interarrival times (a)~in the blue event area and (b)~in the red event area for the \textit{easy} environment, accumulated by the whole swarm. Measurements were taken during a typical run with event rates $\lambda_B=1/(2\times 10^4)$ and $\lambda_R=1/10^4$. We additionally plot the Weibull distribution corresponding to these event rates (``True distribution'') and the Weibull distribution fitted to the measurements.
  (c)~Comparison of number of observations over the whole swarm for: DMMD (standard), communication-free baseline, and DMMD approach with a higher prior;
  lighter colored boxes give performance in \textit{easy} environment, darker boxes give performance in \textit{difficult} environment.}
  \label{fig:weibull_events}
\end{figure*}

As our robots measure an interarrival time between their own arrival to an area and the first observed event, they may systematically underestimate the true interarrival times. 
Fig.~\ref{fig:weibull_events} shows the observed interarrival times of generated events in the blue area (left) and the red area (right) accumulated by the whole swarm during a typical run.
For both areas, the measured interarrival times suffer from underestimating the true interarrival time (and subsequently overestimating the event rate). 
This is caused by two effects.
(1) The robots aim to only observe one event. As a result, they rarely measure the interarrival times between two events but normally between their arrival and the next event.
(2) As robots usually do not arrive at the same times as events, but at a ``later'' time (from the perspective of the Poisson process), they measure shorter interarrival times.
This is represented in the disproportionate representation of short interarrival times.
However, both sides are affected by a similar degree of underestimation and as a result, the relative order of event rates is not affected.
We conclude that our model is appropriate to reach a decision about the relative order of the rates. 
With sufficiently large sample sizes, the observed interarrival times approach the true underlying distribution, but underestimation effects might prevent an accurate estimation at termination time.
For improved estimation of interarrival times, these observations should be treated as left-censored data points; and observation periods without any events should be treated as right-censored events.

\subsection{Accuracy and consensus time}

\begin{figure*}
    \includegraphics[width=\linewidth]{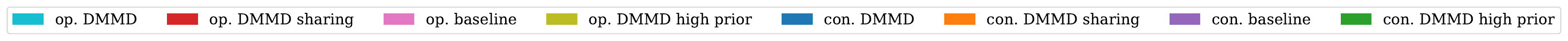}
    \begin{subfigure}[t]{0.32\textwidth}
    \includegraphics[width=\linewidth]{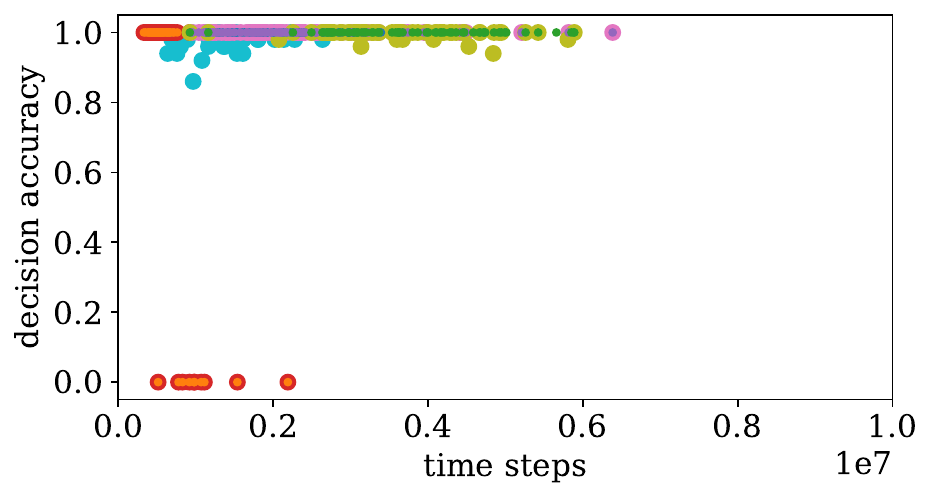}
    \caption{Scatterplot of accuracy and termination times for all algorithms under consideration in the \textit{easy} environment. Each point represents a single experimental run.}
    \label{fig:accuracy_easy}
    \end{subfigure}
    \hfill
    \begin{subfigure}[t]{0.32\textwidth}
    \includegraphics[width=\linewidth]{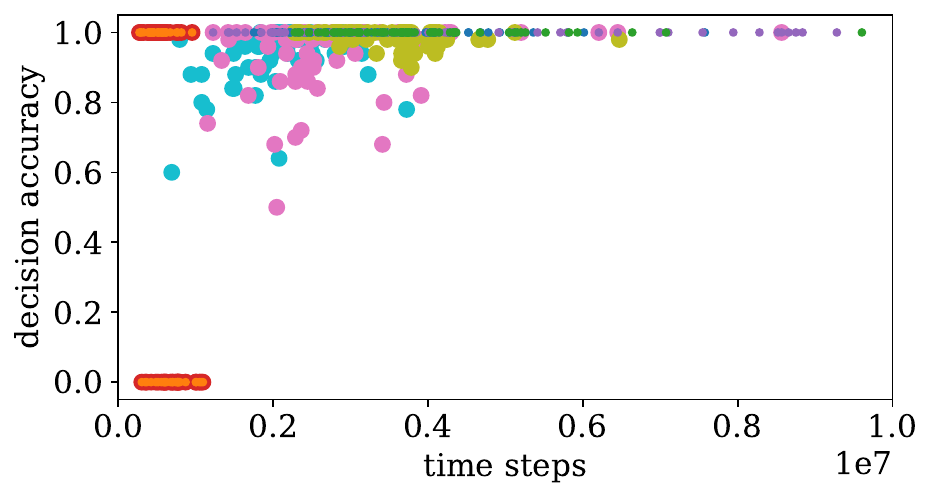}
    \caption{Scatterplot of accuracy and termination times for all algorithms under consideration in the \textit{difficult} environment. Each point represents a single experimental run.}
    \label{fig:accuracy_difficult}
    \end{subfigure}
    \hfill
    \begin{subfigure}[t]{0.32\textwidth}
        \includegraphics[width=\linewidth]{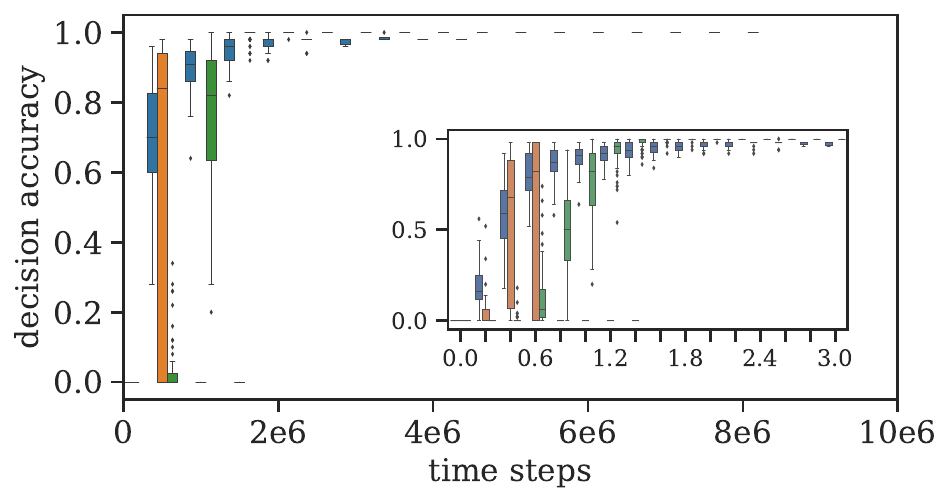}
    \caption{Accuracy over time for consensus-reaching experiments in  \textit{easy} environment; 
    converged runs are excluded after their termination; inset: first $3\times10^6$ time steps.}
    \label{fig:time_progression}
    \end{subfigure}
    \caption{Decision accuracy of the swarm across all experiments. (a) and (b) show the decision accuracy at termination time for the two environments, whereas (c) shows the decision accuracy over time in the \textit{easy} environment.}
\end{figure*}

The key goal in collective decision-making is accuracy, while decision time is another crucial quality measure, reflecting the well-known time–accuracy tradeoff~\cite{valentini2017best}.
Fig.~\ref{fig:accuracy_easy} shows the accuracy and speed of the decision making process in the \textit{easy} environment.
When the termination criterion is set to consensus-reaching (``con.''), all algorithms (except for \textit{DMMD sharing}) always reach the correct decision.
For the termination criterion of opinion-selection (``op.''), we observe that accuracies are generally slightly lower.
This stems from the fact that some robots might initially form a wrong opinion, but will change it at a later point.
This is also reflected in the fact, that the experiments that terminate at opinion-selection finish faster than at consensus (e.g., \num{55} \textit{DMMD} experiments terminate below 2~million time steps under opinion-selection, but only \num{44} experiments terminate under consensus-reaching in the same time).
A~notable exception is the \textit{DMMD sharing} approach.
As expected, the addition of communication leads to a speedup in decision-making.
However, this comes at an increased risk of converging towards the wrong opinion as expected from the speed-vs-accuracy tradeoff. 
In addition, we observe no difference between the consensus-reaching or opinion-selection termination criteria for \textit{DMMD sharing}.
This can be explained by our observation that the additional, shared information strongly influences the internal belief model.
This creates a multiplier effect: recruited robots spread both their opinion and underlying beliefs. However, biased initial observations can quickly lead the swarm to converge on the wrong opinion. 
This is also supported by the time progression of selected opinions (see Fig.~\ref{fig:time_progression}). 
Most runs of the \textit{DMMD sharing} algorithm form complete opinions around $4\times 10^5$ time steps and then quickly reach consensus. 
Conversely, \textit{DMMD} requires approx. $1.5\times 10^6$ time steps before all robots have selected an opinion and the average consensus time is $1.9\times 10^6$ time steps.
A few runs notably achieve high accuracy but require significantly more time steps to reach consensus. This occurs when a robot has made highly biased observations that skew its belief. It must gather additional observations, often until the ring buffer overwrites the outlier, before it converges on the correct decision.



In the case of the \textit{difficult} environment (as seen in Fig.~\ref{fig:accuracy_difficult}), we observe similar trends as for the \textit{easy} environment.
In particular, the consensus-termination leads to correct estimates, except for \textit{DMMD sharing}.
Due to the more similar nature of the rates, it is much more likely that robots are misguided in their initial observations.
In the dissemination phase, the robots then spread their beliefs to other neighbors, effectively recruiting the swarm to the wrong opinion. 
Under the opinion-selection termination criterion, decision accuracy drops, unsurprisingly, as similar event rates increase the chance of misleading observations. Still, extended runs would likely reach the correct consensus. Termination times for \textit{DMMD} and \textit{baseline} are higher than in the \textit{easy} environment. 
For \textit{higher prior}, termination times showed no significant difference. In the \textit{easy} environment, more initial observations were unnecessary and delayed termination. In the \textit{difficult} environment, they improved estimate accuracy, matching \textit{DMMD} performance, which was more affected by poor initial opinions. 
\textit{DMMD} was able to reach consensus correctly in both cases.
Time to consensus was reduced compared to \textit{baseline}. 
Communication improved consensus times, but early biases may result in low accuracy. 
Terminating once all robots selected an opinion improved termination time but reduced accuracy. Still, in the \textit{difficult} environment, most robots held the correct opinion at termination.

\subsection{Observation efficiency}


In the motivation of the scenario, we stated that our goal was to minimize the number of observed events.
Fig.~\ref{fig:observed_events} shows the number of observed events across all experiments.
We excluded the \textit{DMMD sharing} algorithm, as it did not always converge to the correct opinion.
In the \textit{easy} environment, the \textit{DMMD} algorithm minimizes the number of observed events compared to the alternative approaches.
As the \textit{baseline} algorithm does not include any communication, robots need to visit both areas more often to form an opinion.
For the higher prior, the initial belief is a high interarrival time. 
Robots choose longer observation times, potentially observe multiple events instead of one, and hence operate more risky. 
Only after having had multiple observations, their prior is updated to more accurate interarrival times.

For the \textit{difficult} environment with more similar event rates, we do not observe significant differences between the median numbers of observed events.
However, the 3rd quartile in the baseline is much higher than in the two DMMD variants.
This shows that the majority vote can possibly be a guard against worst-case performance when the event areas are similar.
The higher prior shows no significant difference from the standard parametrization, suggesting that early observations yield more reliable samples and thus fewer required visits. 

\section{Discussion and Conclusion}
In our novel collective-decision-making scenario robots need to agree on which of two areas has a lower event rate. 
We propose a Bayesian decision-making framework, where option quality can only be assessed by observing times between random events. 
Our results show that combining the DMMD algorithm with Bayesian inference accurately reaches consensus, while minimizing observed events compared to a baseline. 
Additional communication of robot beliefs leads to a significant number of wrong opinions.
We compared two priors: one underestimating and one overestimating interarrival times. In the \textit{easy} environment, the lower prior outperformed in both termination time and event count. In the \textit{difficult} environment, differences were smaller. A~more systematic study is needed to guide prior selection under uncertain event rates.

Several improvement are possible. Robots tend to underestimate interarrival times, biasing early decisions. Future work could enhance robustness against this effect, especially combined with sharing beliefs.
We could also include explicit modeling of right-censored observations and information-theoretic decisions for sampling to maximize expected information gain. 
Current majority voting can prolong runs if a few robots hold early, biased opinions. Incorporating an exploration phase could reduce runtime by improving estimates~\cite{chin2023minimalistic}, though this would need to be balanced against minimizing observations. Confident robots could further reduce sampling by skipping unnecessary visits.
In addition, we could consider confidence-weighted belief sharing.

We aim to promote more realistic benchmarks in collective decision-making, where assessing option quality is itself a challenge requiring more advanced strategies. Tradeoffs between sampling effort and risk management are relevant for multi-robot applications in environmental monitoring. 
Despite our many simplifications (e.g., no damage/degradation from hazardous events, no noise in observations, no communication failures), we believe this work is a first step toward more realistic benchmarks that also require an intelligent strategy to explore the option qualities themselves.
Future research should explore further challenges, including a more realistic model of hazards and noise, multi-cue option qualities, 
sparse or delayed feedback, and sequential tasks, such as multi-armed bandits over paths.



\newpage
\bibliographystyle{ieeetr} 
\bibliography{demiurge-bib/definitions,demiurge-bib/author,demiurge-bib/address,demiurge-bib/proceedings,demiurge-bib/journal,demiurge-bib/publisher,demiurge-bib/series,demiurge-bib/institution,demiurge-bib/bibliography,references}

\end{document}